%% file: main.tex
\documentclass{article}
\usepackage{microtype}
\usepackage{graphicx}
\usepackage{subfigure}
\usepackage{booktabs} % for professional tables

\usepackage{hyperref}

\usepackage[accepted]{mlsys2022}

\usepackage[newfloat,frozencache,cachedir=.]{minted}
\setminted[python]{frame=lines, tabsize=2, linenos, breaklines, fontsize=\scriptsize}
\usepackage{caption}
\newenvironment{code}{\captionsetup{type=listing}}{}
\SetupFloatingEnvironment{listing}{name=Code}

\usepackage{multirow}
\usepackage{amsmath}

\def\realnumbers{\Re} % TODO: \mathbb{R} didn't work.

\hypersetup{draft} % disable hyperlinks

\usepackage{algorithmic}

\begin{document}

\twocolumn[
\mlsystitle{Learning Compressed Embeddings for On-Device Inference}

% It is OKAY to include author information, even for blind
% submissions: the style file will automatically remove it for you
% unless you've provided the [accepted] option to the mlsys2022
% package.

% List of affiliations: The first argument should be a (short)
% identifier you will use later to specify author affiliations
% Academic affiliations should list Department, University, City, Region, Country
% Industry affiliations should list Company, City, Region, Country

% You can specify symbols, otherwise they are numbered in order.
% Ideally, you should not use this facility. Affiliations will be numbered
% in order of appearance and this is the preferred way.
\mlsyssetsymbol{equal}{*}

\begin{mlsysauthorlist}
	\mlsysauthor{Niketan Pansare}{apple}
	\mlsysauthor{Jay Katukuri}{apple,jpmorgan,equal}
	\mlsysauthor{Aditya Arora}{apple,equal}
	\mlsysauthor{Frank Cipollone}{apple}
	\mlsysauthor{Riyaaz Shaik}{apple}
	\mlsysauthor{Noyan Tokgozoglu}{appleny}
	\mlsysauthor{Chandru Venkataraman}{apple}
\end{mlsysauthorlist}

\mlsysaffiliation{apple}{Apple, Cupertino, CA, USA}
\mlsysaffiliation{appleny}{Apple, New York, NY, USA}
\mlsysaffiliation{jpmorgan}{JP Morgan Chase, New York, NY, USA}

\mlsyscorrespondingauthor{Niketan Pansare}{niketan\_pansare@apple.com}
% You may provide any keywords that you
% find helpful for describing your paper; these are used to populate
% the "keywords" metadata in the PDF but will not be shown in the document
\mlsyskeywords{model compression, neural networks, embeddings, recommendation systems, hardware efficient machine learning methods, efficient model inference}

\vskip 0.3in

\begin{abstract}
In deep learning, embeddings are widely used to represent categorical entities such as words, apps, and movies. An embedding layer maps each entity to a unique vector, causing the layer's memory requirement to be proportional to the number of entities. In the recommendation domain, a given category can have hundreds of thousands of entities, and its embedding layer can take gigabytes of memory. The scale of these networks makes them difficult to deploy in resource constrained environments, such as smartphones. In this paper, we propose a novel approach for reducing the size of an embedding table while still mapping each entity to its own unique embedding. Rather than maintaining the full embedding table, we construct each entity's embedding ``on the fly'' using two separate embedding tables. The first table employs hashing to force multiple entities to share an embedding. The second table contains one trainable weight per entity, allowing the model to distinguish between entities sharing the same embedding. Since these two tables are trained jointly, the network is able to learn a unique embedding per entity, helping it maintain a discriminative capability similar to a model with an uncompressed embedding table. We call this approach MEmCom (Multi-Embedding Compression). We compare with state-of-the-art model compression techniques for multiple problem classes including classification and ranking using datasets from various domains. On four popular recommender system datasets, MEmCom had a 4\% relative loss in nDCG while compressing the input embedding sizes of our recommendation models by 16x, 4x, 12x, and 40x. MEmCom outperforms the state-of-the-art model compression techniques, which achieved 16\%, 6\%, 10\%, and 8\% relative loss in nDCG at the respective compression ratios. Additionally, MEmCom is able to compress the RankNet ranking model by 32x on a dataset with millions of users' interactions with games while incurring only a 1\% relative loss in nDCG.
\end{abstract}
]

% This command actually creates the footnote in the first column
% listing the affiliations and the copyright notice.
% The command takes one argument, which is text to display at the start of the footnote.
% The \mlsysEqualContribution command is standard text for equal contribution.
% Remove it (just {}) if you do not need this facility.

%\printAffiliationsAndNotice{}  % leave blank if no need to mention equal contribution
\printAffiliationsAndNotice{\mlsysEqualContribution} % otherwise use the standard text.

\input{intro}

\input{related-works}

\input{our-approach}

\input{experiments}

\input{conclusions}

%%
%% The acknowledgments section is defined using the "acks" environment
%% (and NOT an unnumbered section). This ensures the proper
%% identification of the section in the article metadata, and the
%% consistent spelling of the heading.
% \begin{acks}
% To Robert, for the bagels and explaining CMYK and color spaces.
% \end{acks}

%%
%% The next two lines define the bibliography style to be used, and
%% the bibliography file.
%\bibliographystyle{ACM-Reference-Format}
\bibliographystyle{mlsys2022}
\bibliography{main}

%%
%% If your work has an appendix, this is the place to put it.
\clearpage
\appendix
\input{appendix}
%\section{Appendix 1}

\end{document}

%% file: intro.tex
\section{Introduction}

In many modern deep learning systems, categorical features with large vocabulary sizes are some of the most important predictive features. Natural language processing systems rely heavily on inputs such as words, sub-word tokens, and individual characters, which are frequently fed directly into the models as categorical features. Additionally, search and recommender systems have increasingly represented model inputs as categorical features, shifting away from the focus on featurizing these inputs using more traditional and use-case specific methods. For example, queries ~\cite{10.1145/2740908.2742774}, documents~\cite{DBLP:journals/corr/abs-1802-03162}, metadata ~\cite{Vasile_2016}, and users have all been treated as categorical features in modern search and recommendation systems. This strategy has proven successful, improving model performance and user experiences.

Research into systems that rely on categorical features have focused heavily on the use of embeddings. Systems like Word2Vec by Mikolov et. al.~\cite{mikolov2013efficient} and fastText by Bojanowski et. al.~\cite{bojanowski2017enriching} are able to learn embeddings that can be used to represent each of the entities in the vocabulary of a given categorical feature. Systems like BERT by Devlin et. al.~\cite{devlin2019bert} take this idea a step further and are able to generate embeddings ``on the fly'' using the entity and the entity's context. While these systems initially targeted natural language processing use cases, the same ideas have been applied broadly. Today, both pre-trained embeddings and embeddings trained from scratch have become common practice in constructing neural networks~\cite{devlin2019bert, peters2018deep}. In fact, these core ideas have become so popular in deep learning that the original embedding papers have garnered tens of thousands of citations. 

The utility of embeddings for dealing with categorical features and adoption of these techniques has accelerated. However, in certain cases, there are important engineering challenges that arise while leveraging embeddings. In particular, as the vocabulary sizes of categorical features grow and as the number of categorical features in a single model grow, the size of the embedding matrices and memory footprint also grow. This problem is particularly salient for search and recommender systems. While natural language vocabularies are often comprised of tens of thousands of words, the vocabularies of queries, documents, and metadata can easily grow into the millions. The resulting embedding matrices are quite large, and some operations on these models can become prohibitive, especially in low resource settings such as smartphones.
Large embedding layers can be an issue even when with server-side inference. For example, an embedding table for a large social network use-case can be on the order of tens of gigabytes and hence can impact the performance of server-side inference~\cite{FB-Recommendation-Architecture}. 
To alleviate this issue, the researchers at Facebook and Twitter proposed \textit{quotient-remainder} trick and \textit{double hashing} respectively to reduce the size of the embedding table~\cite{DBLP:journals/corr/abs-1909-02107, double-hashing}. In section~\ref{sec:experimentalEvaluation}, we show that our approach outperforms both these techniques.

The size of the embedding table is even more important if the recommendation model has multiple embedding tables of comparable sizes or if the inference is done on-device. On-device inference, especially for recommender systems, has three key advantages. First, it has low inference latency (milliseconds rather than tens of milliseconds) thanks to the specialized processors on modern smartphones that can perform on the order of trillions of operations per second. Second, it can be performed in a privacy-friendly manner without requiring user's potentially sensitive data be sent to the servers. Third, it reduces the cost of hosting the recommendation service. 
These considerations are especially important given the impact of recommender systems; 35\% of Amazon's revenue, 23.7\% BestBuy's growth, 75\% Netflix's video consumption and 60\% YouTube's views come from their recommendation system~\cite{XingPersonalizedRecommendationSystem}.

%Because of this challenge, there has been a great deal of research into methods for compressing these embedding matrices, and into model compression in general.

Overall, we make the following contributions:
\begin{itemize}
    \item We propose a novel technique to compress embeddings with minimal loss of accuracy. In section ~\ref{sec:ourApproach}, we highlight three properties satisfied by our technique that make it ideal for compressing the embeddings over multiple tasks and datasets, as opposed to other state-of-the-art techniques.
    \item We validate the utility of our approach by comparing it with the state-of-the-art techniques on two classes of problems, classification and ranking, on five public datasets (Newsgroup, MovieLens, Netflix, Million Songs and Google Local Review). As a bonus, we include the results on two large-scale industry datasets (Games and Arcade) that confirms the findings of the public datasets.
    \item Finally, we evaluate the CPU and memory impact of our technique on Apple iPhone 12 Pro and Google Pixel 2 using popular on-device frameworks, CoreML and TensorFlow Lite respectively.
\end{itemize}

%% file: related-works.tex
\section{Related Works}\label{sec:relatedWorks}

Here we review four broad approaches to reduce the size of the embedding layer. We can reduce the size of the embedding (section ~\ref{sec:sizeOfEmbedding}) or the number of embeddings (section~\ref{sec:numOfEmbeddings}). Additionally, we can reduce the floating point precision (section~\ref{sec:lowerPrecision}) and also consider sparsifying the embeddings (section~\ref{sec:sparseNN}). In this paper, we will primarily focus on the first two approaches. The latter two approaches can be implemented on top of the first two.

\subsection{Embeddings}\label{sec:embeddingRelatedWorks}

In general, our work on compression builds off of the wealth of research that has gone into embeddings over the past decade. This research began with Word2Vec by Mikolov et. al.~\cite{mikolov2013efficient}, which is a method for assigning a static embedding to each word in a vocabulary such that similar words have similar embeddings. 
This method has since become a popular pre-processing step to allow pretrained models to be leveraged for transfer learning from large corpora.
%a popular pre-processing step for categorical features and enables models trained on small amounts of data to leverage the information present in much larger corpora. 
An important step in embedding research came with contextual embeddings built by systems such as BERT ~\cite{devlin2019bert} and ELMo ~\cite{peters2018deep}. These systems are able to incorporate the context in which a feature appears for each individual training or inference example. 
In general, contextual embeddings have been shown to outperform standard embeddings for a wide variety of use cases.

Importantly, while many systems rely on pre-trained contextual or non-contextual embeddings as a static pre-processing step for categorical features, many systems have increasingly allowed for the fine-tuning of embeddings or even for the training of embeddings from scratch to better fit their own use cases.

\subsection{Low-rank approximation}\label{sec:sizeOfEmbedding}
The simplest approach to reduce the size of the embedding layer is to use lower embedding dimensions. Alternatively, we can decompose the embedding matrix $\textbf{E}_{v, e}$ into two smaller matrices $\textbf{U}_{v, h}, \textbf{V}_{h, e}$ where $v = $ vocab size, $e =$ embedding size, $h = $ hidden size, and $h << e$. Lan et. al.~\cite{DBLP:journals/corr/abs-1909-11942} used this technique referred to as low-rank approximation, or factorized embedding parameterization (along with weight sharing), to reduce the number of parameters of BERT-large~\cite{devlin2019bert} by 18x. 
Denil et. al.~\cite{DBLP:journals/corr/DenilSDRF13} showed that the neural networks are overparameterized, and it is possible to predict 95\% of its parameters using this technique. 
% Natalia: \footnote{Denil et. al.~\cite{DBLP:journals/corr/DenilSDRF13} analyzed the MLP models on MNIST and  convnet on CIFAR-10}
Jaderberg et al.~\cite{DBLP:journals/corr/JaderbergVZ14} and Denton et al.~\cite{DBLP:journals/corr/DentonZBLF14} used this technique to compress large convolutional networks for efficient inference with less than a 1\% drop in accuracy. 

\subsection{Weight sharing}\label{sec:numOfEmbeddings}
Since neural networks are overparameterized, it is possible to reduce the size of a layer via weight sharing. 
This paper proposes the sharing of embeddings for inputs in same hash bucket.  

Shi et al.~\cite{DBLP:journals/corr/abs-1909-02107} proposed a similar idea, the \textit{quotient-remainder} trick, for reducing the size of the embedding table. In this approach, the embedding table $\textbf{E}_{v, e}$  is replaced by two embedding tables $\textbf{U}_{m, e}$ and $\textbf{V}_{v/m, \; e}$, and the embedding $\textbf{E}_{i}$ for an item $i$  is approximated using $\textbf{U}_{i \; mod \; m, \; e} \odot \textbf{V}_{k, e}$ here $k$ is the quotient obtained by dividing $i$ by $m$ and $\odot$ denotes element-wise multiplication.

Instead of grouping the input features, HashedNets\cite{hashednets} groups the weights into much smaller hash buckets, i.e. the (virtual) weight $V_{i,j}$ is mapped to $w_{h(i, j)}$ where $h$ is the hash function. The gradient with respect to the mapped weights can be computed using $\frac{\partial \mathcal{L}}{\partial w_k} = \sum_{i,j} \frac{\partial \mathcal{L}}{ \partial V_{i,j}} \frac{ \partial V_{i,j}}{\partial w_k}$. Gong et al.~\cite{Gong14VectorQuantization} proposed clustering the weights using k-means after training. 

Using feature hashing for dimensionality reduction was first proposed by Weinberger et al. for large scale multitask learning~\cite{weinberger09}.
The authors showed that sparsity in the input features minimizes the hash collision and hence makes it more effective.
To reduce the probability of collision even further, Zhang et al. proposed using two hash functions instead of one~\cite{double-hashing}.

\subsection{Lower precision}\label{sec:lowerPrecision}
Most deep learning frameworks use 32-bit floating point representations of parameters during training and 
the parameters are usually quantized before inference for storage and performance reasons.
Since neural networks are shown to be resilient to noise~\cite{Murray94enhancedmlp}, lower precision has been proposed to reduce the size of the neural network as it can be modeled as noise~\cite{han2015deep_compression}.
Vanhoucke et al.~\cite{Vanhoucke} improved the performance of an early neural network for speech recognition using fixed-point (\texttt{int8}) instructions rather than 32-bit floating point instructions on CPUs without loss of accuracy. Krishnamoorthi~\cite{DBLP:journals/corr/abs-1806-08342} evaluated various techniques for quantizing convolutional neural networks with integer weights and activations. Zhu et al~\cite{DBLP:journals/corr/ZhuHMD16} proposed reducing the precision of weights in neural networks to as low as two bits.

\subsection{Sparsification}\label{sec:sparseNN}
Han et al.~\cite{han2015learning, han2015deep_compression} focused on reducing the size (i.e. the number of non-zero parameters) of fully connected layers of AlexNet and VGG-16 networks using connection pruning, where the weights with small magnitudes are set to zero. Liu et al~\cite{Liu_2015_CVPR} reduced the size of convolutional layers of ResNet-152 network by 90\% with just a 2\% loss of accuracy. LeCun~\cite{optimalbraindamage} suggested using the second derivative of the objective function with respect to the parameters for pruning. These methods generate unstructured random connectivity and hence cannot exploit specialized sparse formats or sparse kernels for efficient execution. Structured pruning~\cite{DBLP:journals/corr/abs-1810-05270, DBLP:journals/corr/WenWWCL16} addresses this issue by pruning a channel or a layer.

\subsection{On-Device Inference}\label{sec:ondeviceInference}

Compared to traditional computation paradigms, on-device inference provides several advantages that include improved latency, low communication bandwidth, and better data privacy. On-device inference with deep neural networks is challenging due to compute and memory resource constraints. These constraints are alleviated by methods discussed in recent research works in three directions: framework based, neural net architecture based, and hardware based optimisations.

We ran our performance experiment on iPhone %\footnote{\tiny \url{https://www.apple.com/iphone/}} 
and used a publicly available on-device inference framework called CoreML. %\footnote{{\tiny  \url{https://developer.apple.com/machine-learning/core-ml/}}}. 
CoreML contains direct support to convert models from frameworks like Caffe, TensorFlow, PyTorch to the CoreML format and perform inference using the CPU, GPU, and Neural Engine. Other popular framework for on-device inference are Caffe2~\cite{NEURIPS2019_9015}, TensorFlow Lite~\cite{tensorflow2015-whitepaper}  and, Mobile AI Compute Engine. %\footnote{\tiny  \url{https://github.com/XiaoMi/mace}}.

Neural network architecture research has focused on optimising models for resource constraints on-device by techniques like pruning~\cite{DBLP:journals/corr/abs-1810-05270, DBLP:journals/corr/WenWWCL16, han2015learning, han2015deep_compression, optimalbraindamage}, quantization~\cite{Gong14VectorQuantization}, weight sharing~\cite{DBLP:journals/corr/abs-1909-02107, hashednets} , low rank approximations~\cite{DBLP:journals/corr/abs-1909-11942, devlin2019bert, DBLP:journals/corr/DenilSDRF13, DBLP:journals/corr/JaderbergVZ14, DBLP:journals/corr/DentonZBLF14}, and knowledge distillation~\cite{knowledge-distillation}. These techniques reduce model size and achieve faster computation, with lower memory footprints, thereby reducing inference latency.

Hardware research involves making architectural changes for efficient on-device inference and publishing software development kits (SDKs) to expose the architecture improvements. However these enhancements are mostly vendor specific and cannot be reused across all platforms.

%% file: our-approach.tex
\section{Background}

In neural networks, an embedding layer is typically used to map an input category to a feature vector. 
It does so by using an embedding matrix or table $\textbf{E}^{v \times e}$ where $v$ denotes
the vocabulary size and $e$ denotes the dimensions of the row vector, commonly referred to as the embedding size.  
There are two common ways to implement the embedding layer in a deep learning framework: matrix and table approach. % (shown in the code below).
In the matrix approach, we first map the input to a one-hot vector and then multiply it with the embedding matrix $\textbf{E}$,
whereas in the table approach, no such mapping is required and the embedding vector is obtained via a lookup operation.
%The code below shows an example implementation of these approaches using TensorFlow.

%\begin{minted}{python}
%import tensorflow as tf
%import numpy as np
%v, e = 100000, 128 
%input = [0, 1, 4, 2] # batch size = 4
%E = tf.constant(np.random.rand(v, e), dtype=tf.float32)
%tf.linalg.matmul(tf.one_hot(input, v), E) # Matrix implementation
%tf.nn.embedding_lookup(E, input) # Table implementation
%\end{minted}

The GPU operator for the embedding layer in Caffe\footnote{{\scriptsize \url{https://git.io/JzV9u}}} 
was initially implemented using the matrix approach. Almost all modern deep learning frameworks rely on the table approach for the reasons mentioned below.
Since both approaches require us to store $\textbf{E}$, the storage (on-disk memory) complexity for each approach is $O(v \times e)$.
The runtime memory requirements for the matrix and table approaches are $O(v \times e +  b \times (e+v))$ and 
$O(v \times e +  b \times (e+1))$  respectively, where $b$ is the batch size. Expanding the batch size term shows that the final term in the matrix approach is $O(b \times v)$ as opposed to $O(b)$ in the table approach. In other words, as the vocabulary size increases, the table approach scales in terms of the dimension of the row vector, whereas the matrix approach scales in terms of both the dimension of the row vector and the batch size.

Assuming a vocabulary of 100k (which is common for a recommender system)
and a batch size of 1 (i.e. smallest possible batch size), the memory required for the table approach for the embedding 
size 128 and 256 is 51 MB and 102 MB.
This is acceptable for desktop and workstation CPUs and GPUs  
but not for low-memory phones due to limited disk and memory space.
Shipping such a model reliably to millions of devices with slow network speeds is challenging,
and it becomes even more difficult if the model is trained and updated frequently.
To optimize the memory required for inference, on-device frameworks such as CoreML\footnote{{\tiny  \url{https://developer.apple.com/machine-learning/core-ml/}}} and TensorFlow-Lite~\cite{tensorflow2015-whitepaper} use 
memory-mapped IO (via \texttt{mmap}) rather than loading the entire embedding table into the memory.
Typically, the read performance of \texttt{mmap} is reasonably fast, but writes are much slower.
Thus, the inference time for a memory-mapped embedding layer is negligible, but
training time (typically done via Federated Learning~\cite{DBLP:conf/aistats/McMahanMRHA17}) is much slower.
Nonetheless, the on-disk memory requirements remain the same
and hence models with \textbf{large embedding layers (in hundreds of megabytes)
become prohibitively expensive for on-device inference}.

\section{Our approach}\label{sec:ourApproach}

For a technique to be effective in compressing the embedding layer for NLP and recommender systems, we propose that the compression technique must satisfy three properties:
\begin{enumerate}
	\item The compressed embedding layer should minimize the number of categories that share an embedding vector. An optimal compression technique has capacity to map every category to a \textbf{unique vector}.
	\item If the technique employs composition, the operator should be \textbf{simple} enough so as not to overly constrain the search space. This allows the technique to work well on a  variety of machine learning tasks and datasets.
	\item It should be well-suited even when the categories are distributed non-uniformly. For example: commonly used categories, such as words, movies, and apps, are typically \textbf{power law} distributed.
\end{enumerate}

Before we detail the applicability of above properties for the relevant state-of-the-art techniques, we summarize them in the below table:

\begin{tabular}{p{3.5cm}||p{1cm}|p{1cm}|p{1cm}}
	\hline
	& Unique Vector & Simple Op. & Power-law \\ \hline \hline
	Low-rank approximation & Yes & N/A & No \\
	Quotient-remainder & Yes & No & Yes \\
	Naive hashing & No & N/A & Yes \\
	Double-hashing & No & Yes & Yes \\
	Our approach & Yes & Yes & Yes \\
	\hline
\end{tabular}

Since low-rank approximation techniques, such as factorized embedding parameterization, effectively factorize the embedding matrix, they satisfy the first property but 
ignore the distribution of categories. As a result, these may be less-suited for compressing embedding layers when the 
categories are distributed according to power-law. In our experiments, factorized embedding parameterization performed poorly on all tasks and datasets, except Newsgroup, 
compared to other techniques. 
Similarly, the quotient-remainder trick implements QR decomposition of the embedding matrix, but unlike the
low-rank approximation technique, it can handle category skew. However, we observed that it 
did not perform well in our experiments. We argue that the compositional operator (either concatenation or multiplication of quotient-remainder embeddings)
is relatively complex to generalize over the datasets we tested.

Naive hashing does not guarantee a unique embedding vector for the given category
and has a collision rate of $\frac{v}{m} - 1 + (1 - \frac{1}{m})^v$.
To reduce the collision rate,  Zhang et al. \cite{double-hashing} proposed double hashing,
which has much lower collision rate of $\frac{v}{m^2} - 1 + (1 - \frac{1}{m^2})^v$.
Nonetheless, it still cannot guarantee a unique embedding vector for the given category.
We propose Multi-Embedding Compression, or \texttt{MEmCom} for short,
that allows for a unique vector per category and can handle power-law distributed categories.

Before we discuss \texttt{MEmCom}, let's review the \textit{quotient-remainder} method (described by Algorithm~\ref{alg:quotientRem}) 
proposed by \cite{DBLP:journals/corr/abs-1909-02107}.
The quotient operator is denoted using $\backslash$ and the element-wise multiplication operator is denoted using $\odot$.

\begin{algorithm}
	\textbf{Inputs:} Input category $x$, Embedding tables $\textbf{U} \in \realnumbers^{m \times e}$ and  $\textbf{V} \in \realnumbers^{\frac{v}{m} \times  e}$  \newline
	\textbf{Output:} Embedding vector associated with $x$ \newline
	Determine index $i$ of category $x$  \newline
	Compute hash indices $j = i$ mod $m$ and $k = i \backslash m$ \newline
	Lookup embeddings $\textbf{x}_{rem} = \textbf{U}_{j}$ and $\textbf{x}_{quo} = \textbf{V}_{k}$  \newline
	Return $\textbf{x}_{rem} \odot \textbf{x}_{quo}$
	\caption{Quotient-Remainder Trick~\cite{DBLP:journals/corr/abs-1909-02107}}\label{alg:quotientRem}
\end{algorithm}

Note that the embedding size $e$ is the same for the embedding tables $\textbf{U}$ and $\textbf{V}$,
but the number of embeddings might be different (i.e. $m$ and $\frac{v}{m}$) respectively.
As such, the Quotient-Remainder Method reduces the memory complexity of the embedding layer from $O(v \times e)$
to $O\left((m + \frac{v}{m}) \times e\right)$. 

Logically, we can think of the embedding layer as a set of $v$ functions where $f_v(i) = \textbf{E}_i$.
The hashing method reduces the number of functions to $m$, which can be significantly smaller than that in the uncompressed model (i.e. $v$).
To remedy this, Shi et al.~\cite{DBLP:journals/corr/abs-1909-02107} proposed combining the hashed embeddings (i.e. $\textbf{x}_{rem}$) 
with the quotient embeddings (i.e. $\textbf{x}_{quo}$)  to produce to the final embeddings. 
With this method, the number of functions ($m \times \frac{v}{m}$) are same as that of the uncompressed model (i.e. $v$),
but these functions are constrained rather than arbitrary functions.

Unlike quotient-remainder and double hashing, the embedding tables in \texttt{MEmCom} 
have hybrid shapes and require \textit{broadcasting}.  
Broadcasting is an algorithmic technique that allows matrix/tensor frameworks such 
NumPy, TensorFlow, and PyTorch to handle arithmetic operations (elementwise multiplication in our case) of different shapes~\cite{harris2020array, tensorflow2015-whitepaper, NEURIPS2019_9015, broadcasting}. For example: if we multiply two matrices \textbf{A, B} of shape 3  X  4 and 3 X 1 respectively, we can loop over each column of \textbf{A} and multiply it with \textbf{B}. As looping is typically an expensive operation in these frameworks, broadcasted operators typically have low-level efficient implementation that avoids unnecessary copies of the data.

\begin{table*}[t]
	\begin{tabular}{|p{2cm}|p{4cm}|p{3.75cm}|p{4cm}|} \hline
		& \textbf{Section \ref{sec:rankingExperiment}} & \textbf{Section \ref{sec:classificationExperiment} } & \textbf{Section \ref{sec:inferencePerfExperiment}} \\ \hline
		\textbf{Criteria} & \multicolumn{2}{c|}{Memory (on-disk) v/s accuracy tradeoff} & Inference time and memory footprint \\ \hline
		\textbf{Evaluate} & \multicolumn{2}{c|}{Different model compression techniques} & MEmComp  \\ \hline
		\textbf{Baseline} & \multicolumn{2}{c|}{Uncompressed model} & Weinberger's hashing trick  \\ \hline
		\textbf{Task} & Ranking & \multicolumn{2}{c|}{Classification} \\ \hline
		\textbf{Model} & Point-wise and Pair-wise Ranking Networks & \multicolumn{2}{c|}{Embedding-based Fully connected Feed-forward Network (see code~\ref{code:model})}  \\ \hline
		\textbf{Dataset} (see table \ref{tab:datasets}) & MovieLens, Netflix, MSD, Google, Arcade & Games, Arcade, Newsgroup & MovieLens, Netflix, MSD, Google, Games, Arcade   \\ \hline
		\textbf{Results} & Figure~\ref{fig:plotCommonRanking} and \ref{fig:plot5} & Figure~\ref{fig:plot123} & Table~\ref{tab:runtimePerf} \\ \hline
	\end{tabular} \label{tab:descriptionExperiment}
	\caption{Summary of Experimental Evaluation}
\end{table*}

\texttt{MEmCom} employs a judicious design of the hybrid embedding tables along with easily generalizable linear relationship and a ubiquitous broadcasting operator
to achieve a high compression ratio and low loss of overall accuracy compared to an uncompressed model.
Instead of learning an $e$-dimensional embedding table for quotient, we learn a one-dimensional embedding table which 
can then be composed with the embeddings learned using the hashing method. 
Doing so guarantees that we learn $v$ distinct functions $f_v(i) = \textbf{E}_i$.
We detail this technique in Algorithm~\ref{alg:ourApproachNoBias}.
Since the embeddings $\textbf{U}$ and $\textbf{V}$ are learned jointly,
the embeddings $\textbf{U}$ learned using this technique and that by Algorithm~\ref{alg:quotientRem} 
are likely going to be different. 
Using experimental evaluation, we show that this technique works well over different learning problems.

\begin{algorithm}
	\textbf{Inputs:} Input category $x$, Embedding tables $\textbf{U} \in \realnumbers^{m \times e}$ and  $\textbf{V} \in \realnumbers^{v \times  1}$ \newline
	\textbf{Output:} Embedding vector associated with $x$ \newline
	Determine index $i$ of category $x$ (sorted by frequency)  \newline
	Compute hash index $j = i$ mod $m$ \newline
	Lookup embeddings $\textbf{x}_{rem} = \textbf{U}_{j}$ and $\textbf{x}_{mult} = \textbf{V}_{i}$  \newline
	Return $\textbf{x}_{rem} \odot \textbf{x}_{mult}$ 
	\caption{MEmCom (no bias)}\label{alg:ourApproachNoBias}
\end{algorithm}

We now extend the above algorithm to support bias. 
Unlike the quotient-remainder trick (i.e. Algorithm~\ref{alg:quotientRem}), $\textbf{x}_{mult}$ and $\textbf{x}_{bias}$ are broadcasted in our approach (i.e. Algorithm \ref{alg:ourApproachNoBias} and \ref{alg:ourApproachWithBias}). Though adding $\textbf{x}_{bias}$ allows \texttt{MEmCom} to generalize well on wider variety of tasks and datasets, in practice, \texttt{MEmCom} with no bias performs equally well.

\begin{algorithm}
	\textbf{Inputs:} Input category $x$, Embedding tables $\textbf{U} \in \realnumbers^{m \times e}$,  $\textbf{V} \in \realnumbers^{v \times  1}$ and $\textbf{W} \in \realnumbers^{v \times  1}$ \newline
	\textbf{Output:} Embedding vector associated with $x$ \newline
	Determine index $i$ of category $x$ (sorted by frequency)  \newline
	Compute hash index $j = i$ mod $m$ \newline
	Lookup embeddings $\textbf{x}_{rem} = \textbf{U}_{j}$, $\textbf{x}_{mult} = \textbf{V}_{i}$ and $\textbf{x}_{bias} = \textbf{W}_{i}$ \newline
	Return $\textbf{x}_{rem} \odot \textbf{x}_{mult} + \textbf{x}_{bias} $
	\caption{MEmCom (with bias)}\label{alg:ourApproachWithBias}
\end{algorithm}

\begin{figure*}
	\centering
	\includegraphics[width=\textwidth]{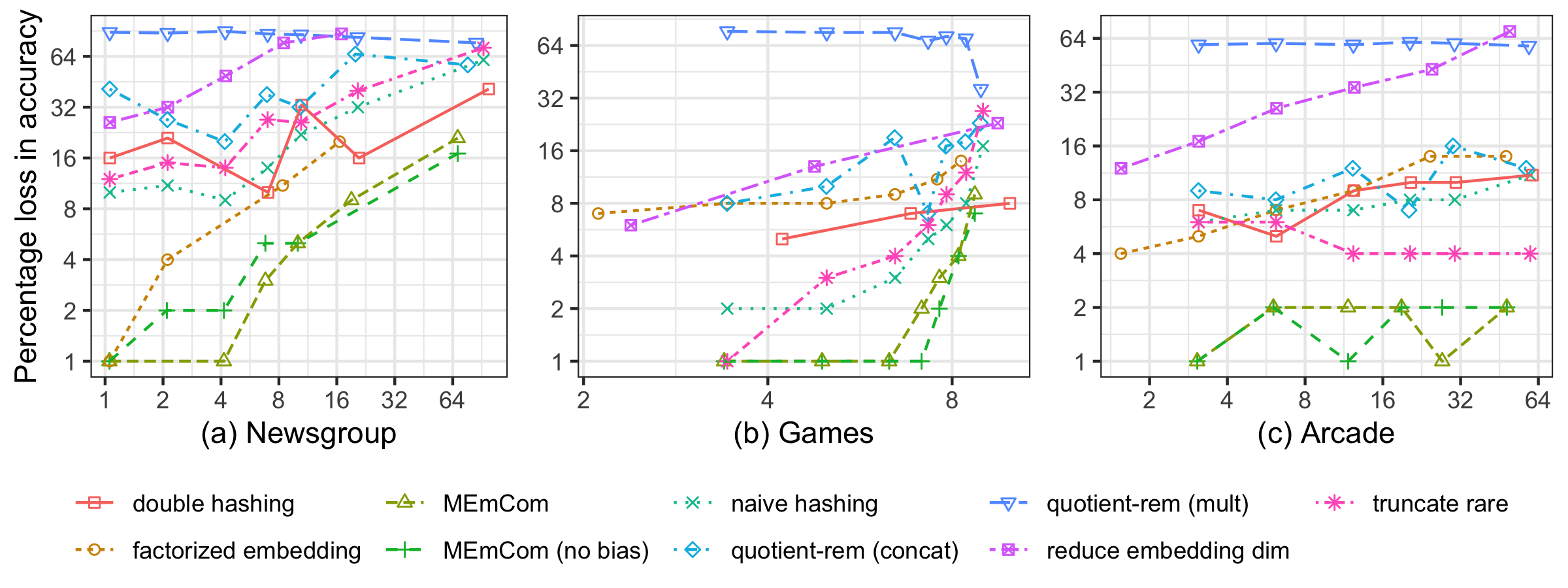}
	\caption{Evaluating compression vs accuracy tradeoff (classification). The x-axis shows the compression ratio.}
	\label{fig:plot123}
\end{figure*}

%% file: experiments.tex
\section{Experimental Evaluation}\label{sec:experimentalEvaluation}

\textbf{Overview.} In this section, we focus on three aspects of the problem that are relevant to on-device inference, 
namely (1) maximum compression of a given model, (2) minimal loss in accuracy,
and (3) efficient on-device performance.
By compression, we refer to reduction in the number of model parameters and hence
the on-disk model size rather than in-memory model size.
Since these models are typically sent to millions of devices (smartphones) over limited bandwidth,
the on-disk model size is important.
%A smaller model will require less disk space and lower network bandwidth to transport the model.
That said, it is possible that a small model on-disk can potentially have a high inference overhead,
i.e. it might require a large amount of RAM or take longer to perform inference
(such as Weinberger's hashing method~\cite{weinberger09}).
To explore these aspects, we have two sets of experiments:
one where we evaluate the tradeoff between model size and model accuracy (see section~\ref{sec:classificationExperiment} and ~\ref{sec:rankingExperiment}))
and another where we focus on the inference overhead of \texttt{MEmCom} (our approach) (see section~\ref{sec:inferencePerfExperiment}).
For the first set of experiments, we compare our technique to the state-of-the art techniques to compress embeddings for two tasks: classification and ranking. These are discussed in sections ~\ref{sec:classificationExperiment} and ~\ref{sec:rankingExperiment} respectively.

\textbf{Model.} To be consistent across the experiments, we use a common network structure and hyper-parameters:
an embedding-based, fully connected, feed-forward network. The network structure is a common strategy for learning latent user embeddings based on items that the user has interacted with in some way. For example, Nazari et al.~\cite{Nazari_2020} used a similar architecture to perform podcast recommendations. Importantly, we observed optimal model performance using this network structure, having experimented with a variety of popular network structures and hyperparameters. The code~\ref{code:model}  
shows the Keras implementation of the baseline network for the classification experiment (section~\ref{sec:classificationExperiment}) where the embedding size is 256, the input length is 128, and the number of embeddings is equal to the size of the input vocabulary \texttt{V}.

\begin{code}
	\begin{minted}{python}
	embed = Embedding(input_dim=V, output_dim=256, input_length=128, mask_zero=False)(input)
	l = AveragePooling1D(pool_size=128)(embed)
	l = Flatten()(l)
	l = ReLU()(l)
	l = Dropout(dropout)(l)
	l = BatchNormalization()(l)
	l = Dense(units=int(embedding_size / 2), activation='relu')(l)
	l = Dropout(dropout)(l)
	l = BatchNormalization()(l)
	output_layer = Dense(units=num_labels, activation='softmax')(l)
	\end{minted}
	\captionof{listing}{Embedding-based Fully connected Feed-forward Network}
	\label{code:model}
\end{code}

Except for the embedding layer (line 1), we use the same network for all the techniques in this experiment.
For the ranking experiments (section~\ref{sec:rankingExperiment}), we train the pointwise network and then use the softmax scores as the basis for ranking. We also alter the network by removing the Dense layer following the Average Pooling, as this empirically proved to give us better performance. Additionally, we experiment with a pairwise siamese network where the above network is shared for the inputs. We explain this in more detail in the section~\ref{sec:rankingExperiment}.

\begin{figure*}
	\centering
	\includegraphics[width=0.9\textwidth]{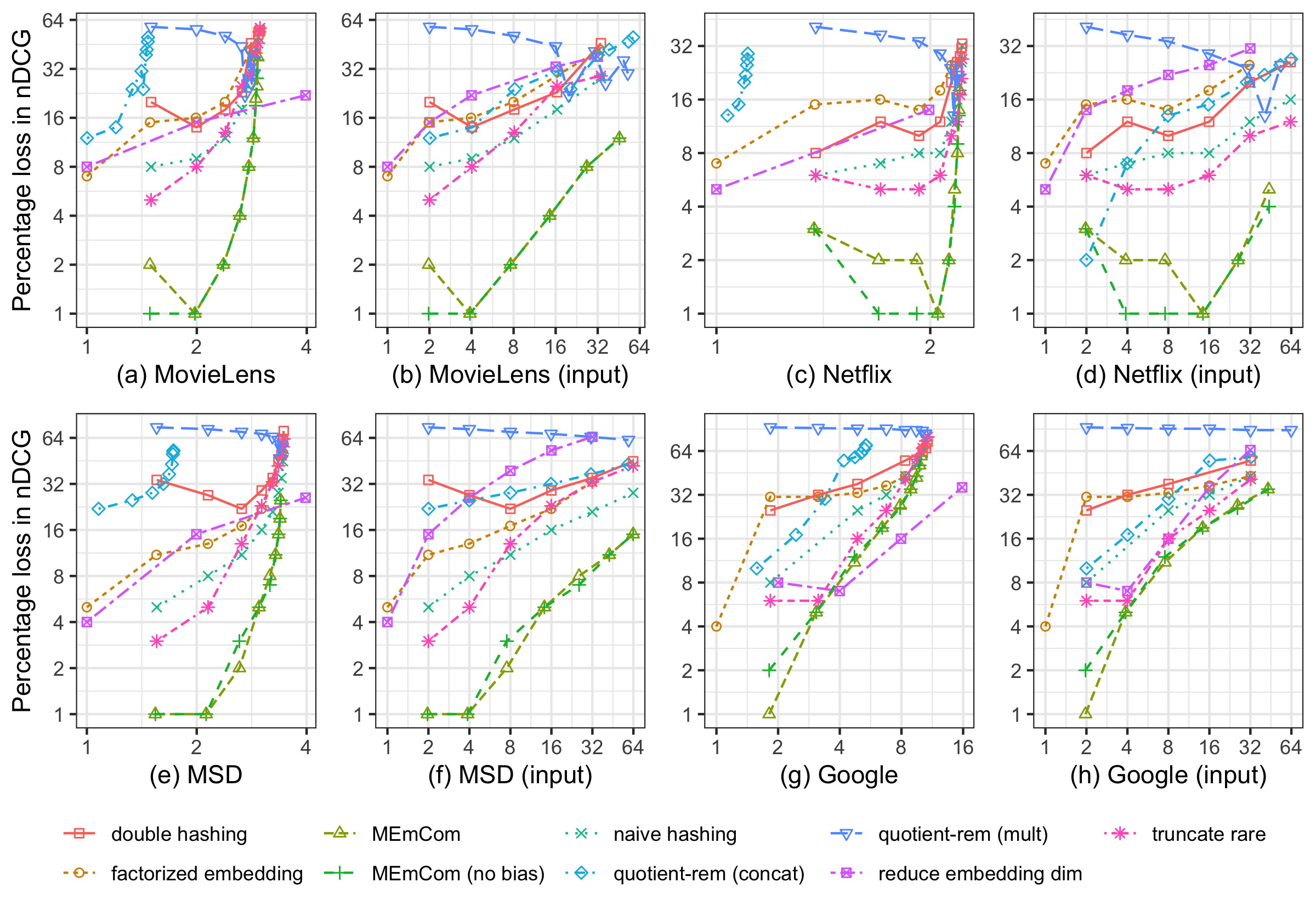}
	\caption{Evaluating compression vs accuracy tradeoff (Pointwise Ranking). The x-axis shows the compression ratio.}
	\label{fig:plotCommonRanking}
\end{figure*}

\textbf{State-of-the-art techniques.} As described in section~\ref{sec:relatedWorks}, we can group the techniques to compress embeddings into two categories, one that reduces the number of embeddings and one that reduces the embedding dimension.
We first discuss the techniques that fall into the first category.
(1) Naive hashing performs the \texttt{mod} operation on the input feature vector before the embedding table lookup.
(2) In double hashing, we use two hash functions (and embedding layers) instead of one and concatenate their embeddings~\cite{double-hashing}.
(3) We use two variants of the \textit{quotient-remainder} method~\cite{DBLP:journals/corr/abs-1909-02107} described in 
the section \ref{sec:numOfEmbeddings}, one where the compositional operator is \textit{concatenation}
and other where the compositional operator is \textit{elementwise multiplication}.
(4) Also, we compare these with algorithm \ref{alg:ourApproachNoBias} (``\textit{MEmCom (our approach with no bias)}'') and 
(5) algorithm \ref{alg:ourApproachWithBias} (``\textit{MEmCom} (our approach)'').
To validate that the hashing-based compression techniques yield useful results, we include a simple baseline where we drop the less popular apps (``\textit{truncate rare}'').
There are two techniques we evaluate that reduce the embedding dimension, namely
\textit{factorized embedding parameterization}~\cite{DBLP:journals/corr/abs-1909-11942} and by simply
\textit{reducing the number of embedding dimensions}. (6) For all methods except ``reduce embedding dim,'' the output embedding dimensions
is 256. For ``reduce embedding dim,'' we progressively reduce the embedding dimensions by a factor of 2 (i.e. 128, 64, 32, 16, 8 and 4)
for compressing the model. Each point on the below plots indicates a model trained using one of these hyperparameters.
(7) For ``factorized embedding,'' we keep the output embedding dimension the same (i.e. the number of hidden units of dense layer = 256), but vary 
the dimension of the embedding layer by  a factor of 2 starting from 128.
For all other approaches, we vary the number of embeddings using the mod $m$ to vocabulary size, 100K, 50K, 25K, 10K, 5K and 1K.
The results for  \texttt{TT-Rec}~\cite{DBLP:journals/corr/abs-2101-11714} were similar to ``factorized embedding''  for all datasets;
likely because both these approaches have large number of shared parameters, which in turn decreases the representational capacity of the embeddings.
\textit{Mixed dimension embeddings}~\cite{DBLP:journals/corr/abs-1909-11810} is a blocked extension of ``factorized embedding'' with two additional
hyperparameters, i.e., the number of blocks and the temperature (in case of popularity-based dimension sizing).
In line with author's suggested rule of thumb, we set the number of blocks to the number of distinct categorical features, which in our case is 1.
With this hyperparameter setting, the results were similar to that of the ``factorized embedding'' approach.

\begin{table*}[t]
	\begin{tabular}{|p{4.3cm}|p{1.5cm}|p{1.5cm}|p{1cm}|p{1.8cm}|p{1cm}|p{1cm}|p{1.5cm}|} \hline
		& Newsgroup & MovieLens & Million Songs & Google Local Reviews & Netflix & Games & Arcade \\ \hline
		Number of training samples & 11.3K & 655K & 4.5M & 246K & 2.1M & 78M & 7.5M \\
		Number of evaluation samples & 7.5K & 72.8K & 500K & 27K & 235K & 65K & 65K \\
		Input vocabulary size & 105K & 10K & 50K & 200K & 17K & 480K & 300K \\
		Output vocabulary size & 20 & 5K & 20K & 20K & 16K & 119K & 145 \\ \hline
	\end{tabular}
	\caption{Datasets Used.}
	\label{tab:datasets}
\end{table*}

\textbf{Datasets.} For all the datasets, the input is a fixed length vector of size 128.
(1) The ``Newsgroup'' dataset is fetched using scikit-learn's dataset API \texttt{sklearn.datasets.fetch\_20newsgroups}.
(2) The ``MovieLens Ratings''~\cite{movielens} dataset is a popular dataset consisting of 25M user ratings of movies, often used for benchmarking recommender systems.  
(3) The ``Million Songs''~\cite{Bertin-Mahieux2011} dataset contains around 1M triplets of the form \textit{user, song, number of listens}. This dataset is commonly used for benchmarking recommender systems as well.
(4) The ``Google Local Reviews''~\cite{He_2017} dataset contains around 11M reviews of local businesses, along with locations and other metadata.
(5) The ``Netflix Ratings''~\cite{inproceedings} dataset refers to the movie ratings dataset released by Netflix for their recommender system competition.
In addition to the above publicly available datasets, we include two proprietary datasets: \textit{Games} and \textit{Arcade}, a random sample of mobile gaming platform users. The Games dataset is about 10x larger than the Arcade dataset and has a much larger output vocabulary. Even though these two datasets are not publicly available, we decided to include the corresponding experiments as they closely reflect the scale and complexity (especially in terms of behavioral signals) of a real-life large-scale recommendation dataset.
The key differences between the above datasets are the lengths of input/output vocabularies
and the distributions of the vocabularies (especially the output vocabularies). Each output vocabulary indirectly affects the number of parameters in the last layer of (see line 10 in the above code)
and hence the size of each model. The different distribution of each output vocabulary provides diverse workloads for this experiment.
We summarize the statistics of these datasets in the table~\ref{tab:datasets}.

\textbf{Setup.} We train our model using TensorFlow 1.12.3 and Keras 2.2.4 on a machine with a Nvidia V100 GPU
and CUDA 9.0 and CuDNN 7.6.2. 

\subsection{Experiment 1: Compression vs. Accuracy tradeoff (Classification)}\label{sec:classificationExperiment}

\begin{figure}
	\centering
	\includegraphics[width=0.45\textwidth]{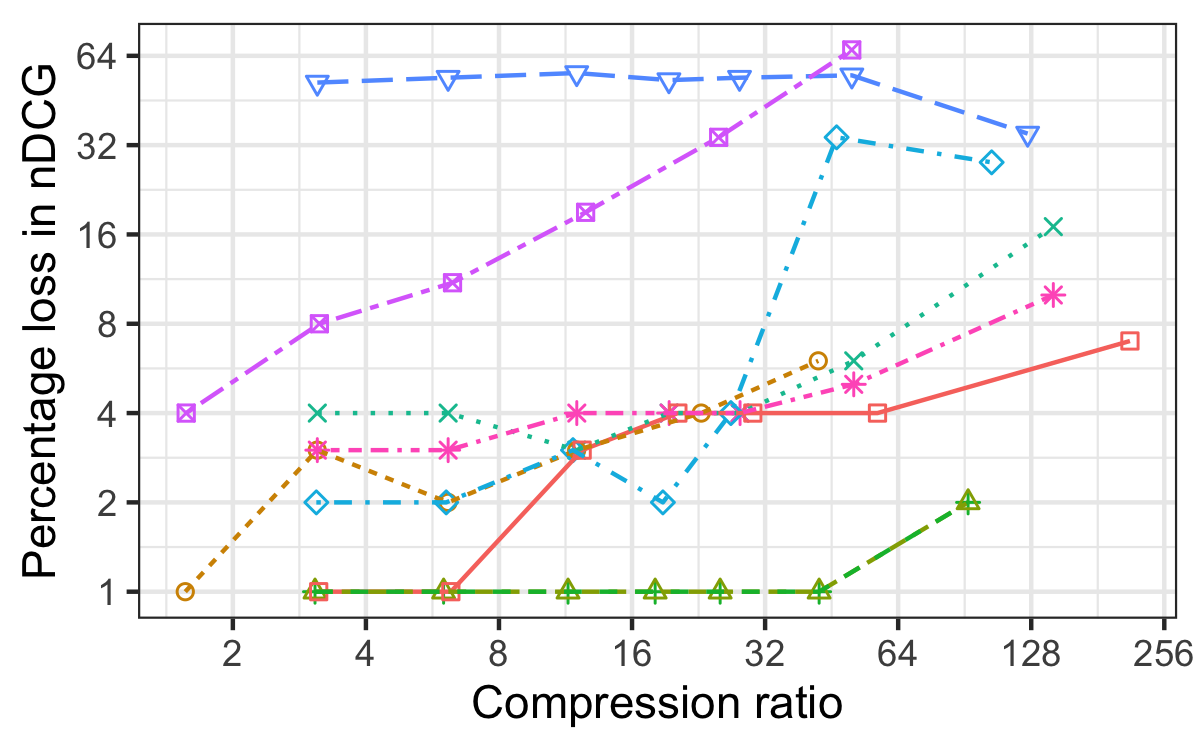} 
	\caption{Evaluating compression vs accuracy tradeoff (Arcade - Pairwise Ranking). This figure has the same legend as the figure~\ref{fig:plotCommonRanking} .}
	\label{fig:plot5}
\end{figure}

\textbf{Experimental Setup.} For each technique, we evaluate the loss of accuracy (compared to the baseline uncompressed network) and the compression ratio (size of baseline network $/$ size of the compressed network)
over three datasets: (1) 20 Newsgroups, (2) Games, and (3) Arcade. 
For all three datasets, the input vocabulary is of size 100K or greater.
For the last two datasets, we use the previous 127 apps that the user purchased along with the user's country to predict the last app the user purchased.
To simplify the setup, we use a shared vocabulary for the app identifiers and the countries.
For example, if there are $n$ countries and $m$ apps, then the vocabulary is of size $n+m+1$.
The countries are mapped to ids $1$ to $n$ and the apps are mapped to ids $n+1$ to $n+m$.
The id $0$ is reserved for padding.
We used frequency-based mapping for the vocabulary, i.e. the most downloaded app is assigned the id $n+1$
and the country with most purchases is assigned the id $1$.
To ensure a fixed length feature vector, we drop the least recently purchased apps if the user has more than 
127 apps and pad (with id $0$) if the user has less than 127 purchases.

\textbf{Results.} Figure~\ref{fig:plot123} shows the state-of-the-art techniques we explored for reducing the size of the embedding layer
for Newsgroup, Games, and Arcade datasets.
The x-axis shows the compression ratio, which is the ratio of the number of parameters of the uncompressed model
and that of the technique for a given set of hyperparameters. 
For consistency across the datasets, we measure the number of parameters of all the layers and not just the embedding layers.
Thus, the compression ratio of 2 implies that the proposed technique was able to
compress the on-disk model size by half, but the embedding layer was compressed even further.
The y-axis shows the percentage loss in accuracy for the given technique compared to the uncompressed model.
For all compression ratios, \texttt{MEmCom} has much lower loss in accuracy compared to other techniques.
Other than \texttt{MEmCom} and factorized embedding parameterization, all other techniques did not work well for the Newsgroup dataset.
On the Arcade dataset, a dumb compression technique ``truncate rare'' where we drop the rare apps, worked pretty well compared to more sophisticated, state-of-art  compression techniques. Even in this case, \texttt{MEmCom} outperformed it by 2x.

\subsection{Experiment 2: Compression vs nDCG tradeoff (Ranking)}\label{sec:rankingExperiment}

\textbf{Experimental Setup.} For the ranking experiments, we worked with five datasets: (1) MovieLens Ratings~\cite{movielens}, (2) Million Songs~\cite{Bertin-Mahieux2011}, (3) Google Local Reviews~\cite{He_2017}, (4) Netflix Ratings~\cite{inproceedings}, and (5) Arcade.
While processing the Movielens Ratings, Google Local Reviews, Million Songs, and Netflix Ratings datasets, we produce up to five training examples per user. The training label for each example will come from the set of the most recent item interactions, with the 128 most recent item interactions (excepting the label) serving as the training input data. Additionally, we filtered items that did not have sufficient popularity in the training data. This decision allows sufficient data to train embeddings that perform well and incents the models to be as efficient as possible with the allowed number of parameters, allowing us to more easily compare different compression techniques. The table in ~\ref{sec:classificationExperiment} provides information on the number of items allowed in the input and the output of the models for each dataset,
For the network architecture, we chose to use the same architecture described by the Keras code in the introduction to this section, setting up a pointwise learning-to-rank network. We altered the network by removing the Dense layer following the Average Pooling, as this proved empirically to give us better performance.

\begin{table*}[t]
	\centering
	\begin{tabular}{|p{2cm}|p{1.8cm}|p{1cm}|p{1cm}|p{1.2cm}|p{1.2cm}||p{1cm}|p{1cm}|p{1.2cm}|p{1.2cm}|} 
		\cline{1-10}
		\multicolumn{2}{|c|}{\multirow{3}{*}{}}     & \multicolumn{4}{c|}{\textbf{Inference Time}} & \multicolumn{4}{c|}{\textbf{Memory footprint}}   \\ 
		\cline{3-10}
		\multicolumn{2}{|c|}{}     & \multicolumn{3}{c|}{CoreML} & TF Lite & \multicolumn{3}{c|}{CoreML} & TF Lite  \\ 
		\cline{3-10}
		\multicolumn{2}{|c|}{}                                    & {\scriptsize \texttt{all} }  & {\scriptsize \texttt{cpuOnly}} & {\scriptsize \texttt{cpuAndGPU}}  & CPU             & {\scriptsize \texttt{all}} & {\scriptsize \texttt{cpuOnly}} & {\scriptsize \texttt{cpuAndGPU}}   & CPU              \\ 
		\cline{1-10}
		\multirow{2}{*}{Newsgroup}           & MEmCom   & 0.21 & 0.38    & 0.47       & 0.18            &   3.23 & 2.88 & 4.88 & 1.55 \\ 
		\cline{2-10}
		& Weinberger & 0.89 & 0.9     & 0.95       & 30.96           &  27.57 & 27.62 & 28.29 & 8.21 \\ 
		\cline{1-10}
		\multirow{2}{*}{Movielens}           & MEmCom   & 0.07 & 0.06    & 0.13       & 0.05            &   2.6 & 2.8 & 4.56 & 1.04 \\ 
		\cline{2-10}
		& Weinberger & 0.9  & 0.91    & 0.95       & 30.84           &  27.6 & 27.6 & 28.4 & 8.21 \\ 
		\cline{1-10}
		\multirow{2}{*}{Million Songs}       & MEmCom   & 0.07 & 0.06    & 0.12       & 0.07            &  2.7 & 2.49 & 4.34 & 1.24 \\ 
		\cline{2-10}
		& Weinberger & 0.91 & 0.9     & 0.96       & 30.6            &   27.8 & 27.9 & 28.5 & 8.22 \\ 
		\cline{1-10}
		\multirow{2}{2cm}{Google Local Review} & MEmCom   & 3.49 & 3.34    & 3.42       & 0.4             &   5.34 & 4.30 & 5.83 & 3.44  \\ 
		\cline{2-10}
		& Weinberger & 1.19 & 1.2     & 1.25       & 30.91           &   10 & 31.7 & 32.6 & 9.25  \\ 
		\cline{1-10}
		\multirow{2}{*}{Netflix}             & MEmCom   & 1.22 & 0.60    & 0.76       & 1.22            &   8.65 &  2.64 & 4.24 & 8.6  \\ 
		\cline{2-10}
		& Weinberger & 1.32 & 1.32    & 1.42       & 31.4            &  10.6 &  37.8 & 38.5 &  31.4 \\ 
		\cline{1-10}
		\multirow{2}{*}{Games}               & MEmCom   & 3.42 & 3.33    & 3.42       & 4.4             &   5.39 & 4.22 & 5.81 & 31.2  \\ 
		\cline{2-10}
		& Weinberger & 2.51 & 2.53    & 2.64       & 34.6            &  13.2 & 16.2 & 16.2  & 37.5  \\ 
		\cline{1-10}
		\multirow{2}{*}{Arcade}              & MEmCom   & 0.06 & 0.06    & 0.12       & 0.01            &   3.63 & 2.52 & 4.38 & 1.36 \\ 
		\cline{2-10}
		& Weinberger & 1.14 & 1.15    & 1.18       & 30.9            &   10.2 & 29.1 & 30 & 37.6   \\
		\cline{1-10}
	\end{tabular}
	\caption{Inference time (in milliseconds) and memory footprint (in megabytes) on different devices (batch size = 1, FP32) comparing MEmCom (no bias) and Weinberger Hashing}
	\label{tab:runtimePerf}
\end{table*}

Due to our training data design, we use the softmax as our loss function as in the classification experiments. After training, we predict on the evaluation data using the softmax scores to compute the final ranking score for each item. This simplified architecture allows us to focus on the impact of the compression techniques we are exploring. 
For the Arcade dataset, instead of using a pointwise ranking model, 
we use a pairwise ranking model with the \textit{RankNet}~\cite{ranknet} architecture (for greater coverage of different types of models in the experiments).
This network takes as input user features and two item IDs such that the first item is ranked higher than the second item.
It outputs two scores corresponding to the input item ids, and  during training, we maximize the difference between these scores.
If the network is able to differentiate between arbitrary pairs of item ids in the evaluation dataset,
it has learned the ranking function and then can be used to rank any list of items available in the output vocabulary.
We use the Arcade dataset described earlier and use this network to rank the Arcade games.
We evaluate a popular ranking metric, normalized discounted cumulative gain (or nDCG for short)~\cite{ndcg}, for evaluation.

\textbf{Results.} Figures~\ref{fig:plotCommonRanking} and ~\ref{fig:plot5} show the percentage loss in nDCG for the techniques we evaluated for reducing the embedding
layer compared to an uncompressed model. Figure~\ref{fig:plotCommonRanking} shows the results of a pointwise learning-to-rank network on the above mentioned dataset, while figure~\ref{fig:plot5} shows the results of the pairwise ranking model on the Arcade dataset.
Like earlier experiments, we prefer techniques that have minimal loss in nDCG
for a given compression ratio. As shown in figure~\ref{fig:plot5}, \texttt{MEmCom} has less than 1\% loss in nDCG 
while compressing the Arcade ranking model by 32x. 
Additionally, we achieved an approximately 4\% loss in nDCG while compressing the input embedding matrices of the Movielens Ratings, Google Local Reviews, Million Songs, and Netflix Ratings models by 16x, 4x, 12x, and 40x, respectively, beating out other state-of-the-art model compression techniques. These results are shown in figures ~\ref{fig:plotCommonRanking} (a), ~\ref{fig:plotCommonRanking} (c), ~\ref{fig:plotCommonRanking} (e) and ~\ref{fig:plotCommonRanking} (g), along with the compression ratios of the full models.  
As \texttt{MEmCom} with and without bias performs exactly the same, their lines overlap in the figure\ref{fig:plot5}.

\subsection{Experiment 3: Runtime Performance of On-Device Inference} \label{sec:inferencePerfExperiment}

\textbf{Experimental Setup.} 
In this experiment we compare the runtime performance of a model that uses our approach described in the section~\ref{sec:classificationExperiment} with a model that applies Weinberger’s hashing method~\cite{weinberger09} on the one-hot encoded input features. To keep the comparison fair, both models have the same set of layers except for the first embedding layer and also the same fixed hash size of 10K is used in both models.
Because the multi-embedding approach, the quotient trick and double hashing all rely on ``hashing the number of embeddings,'' these results 
are applicable for those approaches too.
The two models were benchmarked with the datasets discussed in the sections ~\ref{sec:classificationExperiment} and ~\ref{sec:rankingExperiment}. 
The benchmarks were performed on two smartphones, an Apple iPhone 12 Pro %\footnote{{\tiny\url{https://www.apple.com/iphone-12-pro/}}} 
and a Google Pixel 2, %\footnote{{\tiny\url{https://en.wikipedia.org/wiki/Pixel_2}}}
using the on-device frameworks, CoreML 4.1.4, and TensorFlow-Lite 2.3.0 (TF-Lite) respectively. We report the average values across 1000 benchmark runs.
Since the smartphones are from different generations, the goal is not to compare their performance, nor the performance of the corresponding
on-device frameworks. Instead, we focus on comparing \texttt{MEmCom} with Weinberger’s hashing method for a given setup.
CoreML allows a developer to constrain the possible set of computes using an enum \texttt{MLComputeUnits}, which can be set to \texttt{all}, \texttt{cpuOnly} or \texttt{cpuAndGPU}.
But with the exception of the CPU,  the CoreML API does not allow the developer to force the execution of the model on a specific compute unit.
For example, if the enum is set to \texttt{all}, CoreML can schedule an operator in the model (or the entire model) on the best compute unit available, which can be Neural Engine, CPU or GPU.
TF-Lite's benchmarking tool %\footnote{ {\scriptsize \url{https://www.tensorflow.org/lite/performance/measurement}}} 
was used to measure the performance of the corresponding models.
This tool allows the measurement of execution on a given compute using the flag \texttt{use\_gpu}.
For GPU execution, TF-Lite tries to delegate ineligible operators (one-hot operator) on CPU and schedules the remaining on the GPU.
However, our execution fails as one of the operators (\texttt{reduce\_sum}) scheduled to be executed on GPU does not have a corresponding implementation. %\footnote{{\scriptsize\url{https://www.tensorflow.org/lite/guide/ops_compatibility}}}. 
Hence, we do not include the results for GPU execution with TF-Lite.
The models were not quantized during compilation; the parameters are stored in 32-bit precision and the computation is also performed in the same precision.
After model initialization, the inference time for baseline uncompressed model is comparable to \texttt{MEmCom}. The on device training time will be much lower for \texttt{MEmCom} compared to non-compressed baseline as it has much lower number of parameters and hence much smaller gradients during back-propagation. Initialization and compilation overhead are not included in the results.

\textbf{Results.}
Table ~\ref{tab:runtimePerf} shows inference time (in milliseconds) and the runtime memory footprint (in megabytes) for the model using Weinberger’s hashing trick and the models using our approach. 
\texttt{MEmCom} outperforms Weinberger’s hashing trick for all computes on both smartphones.
This is because our approach uses an efficient lookup operator described in section 3 and Weinberger’s hashing method relies on the one-hot encoded representation.
As CoreML and TF-Lite implement the lookup operator in the embedding layer using \texttt{mmap},
the memory footprint for \texttt{MEmCom} is very small compared to the Weinberger’s hashing method. TF-Lite's \texttt{mmap} is tuned for lower memory footprint than for faster inference time. So we see a considerable difference in memory footprint, inference time betweeen CoreML and TF-Lite.
The above results show that \texttt{MEmCom} has a very low-memory footprint and reasonably small inference time and hence is
well suited to be deployed on off-the-shelf phones.

%% file: conclusions.tex
\section{Conclusions}

In this paper, we propose a novel method for compressing the embedding layer of a neural network without significant loss in the overall accuracy of the model.
Unlike the hashing trick and similar methods, our method allows the network to learn a unique embedding vector for each categorical entity, giving the model an edge in the compression of large-scale search and recommendation models.
We compared our approach with state-of-the-art model compression techniques
on different problem classes and on multiple datasets.
Our experiments validate that our approach significantly outperforms 
other techniques in terms of compression vs. accuracy tradeoffs.
Furthermore, we benchmark the runtime performance of our approach with a state-of-the-art feature hashing technique, on an Apple iPhone 12 Pro and a Google Pixel 2 using popular libraries (CoreML and TensorFlow Lite).  The results shows that our approach is well-suited for on-device inference; thereby paving the way for more privacy-friendly recommendation ecosystems.

%% file: appendix.tex
\section{Additional Experimental Evaluation}

The section~\ref{sec:experimentalEvaluation} focused on evaluating the effectiveness of \texttt{MEmCom} for a typical on-device inference scenario.  In this section, we expand on that to cover more advanced scenarios.

\subsection{Fixed model size} \label{sec:hyperparams}

\textbf{Experimental Setup.} In the section~\ref{sec:experimentalEvaluation}, we assumed that the hyperparameters of the baseline uncompressed model was tuned to achieve best generalization and the goal was to compress the model as much as possible with minimal loss of accuracy. To that end, we evaluated \texttt{MEmCom} under different number of embeddings (i.e. hash sizes), while assuming that the embedding size was the model hyperparameter. This is suitable for most cases where the data scientist chooses a reasonably large embedding size; which in our case was 256. In this experiment, we fix the model size and vary both the embedding size as well as the number of embeddings. This experiment evaluates a scenario where a size budget is imposed by the use-case and the goal is to find the best set of hyperparameters that satisfy that budget. Also, this experiment helps us understand the tradeoff between chosing a large embedding size (and a small number of embeddings) and a large number of embeddings (and a small embedding size) for \texttt{MEmCom}. For the public datasets, we fixed the size of each model to half the size of the corresponding baseline model. For the Arcade and Games datasets, we fixed the model size to be 20MB. Other than that, the experimental setup is same as the one described in the section~\ref{sec:classificationExperiment}.

If we increase the number of embeddings, we have to decrease the embedding size accordingly to ensure that model size remains the same. As the model size also depends on the output vocabulary size, we performed a simple binary search to find the embedding size for corresponding number of embeddings. 

\begin{figure}
	\centering
	\includegraphics[width=0.45\textwidth]{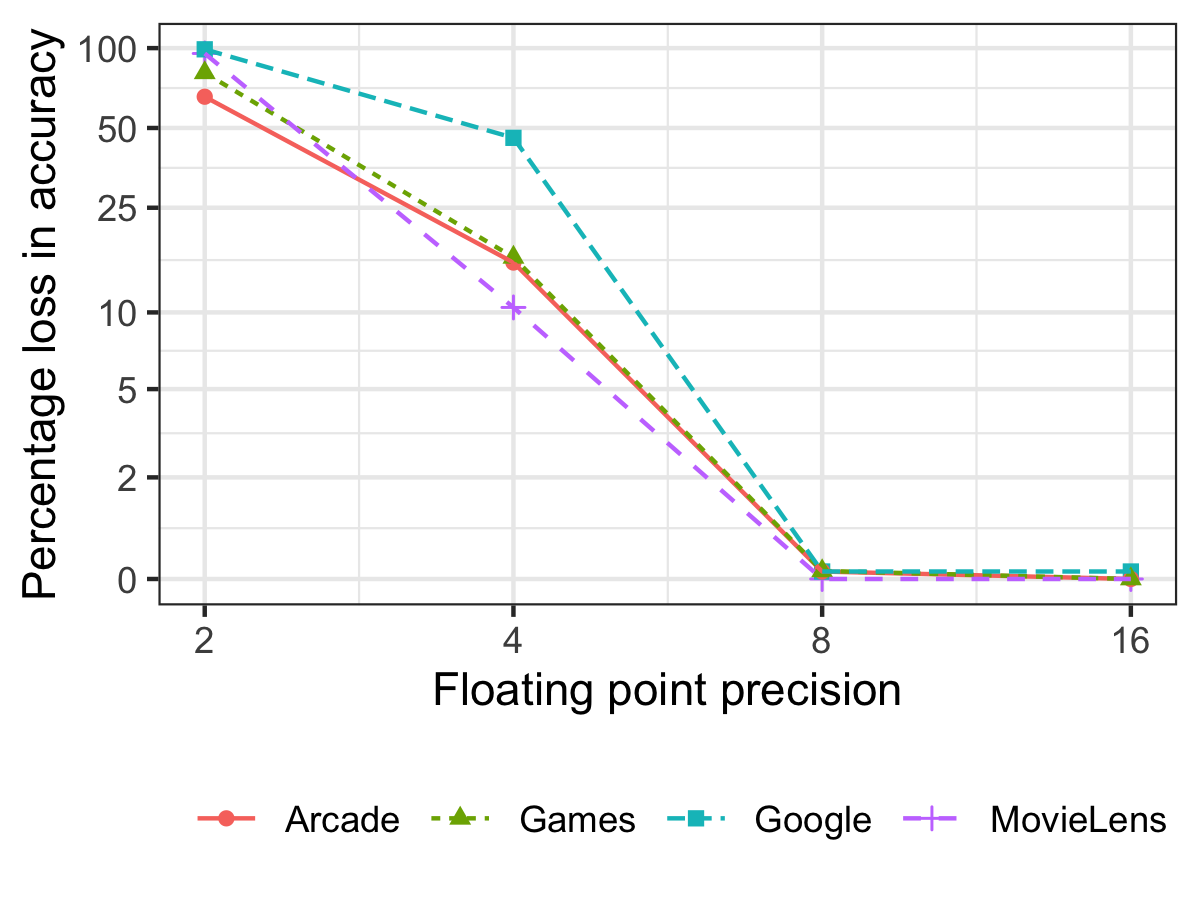} 
	\caption{Accuracy vs floating point precision tradeoff.}
	\label{fig:quantizationPlot}
\end{figure}

\begin{figure*}
	\centering
	\includegraphics[width=\textwidth]{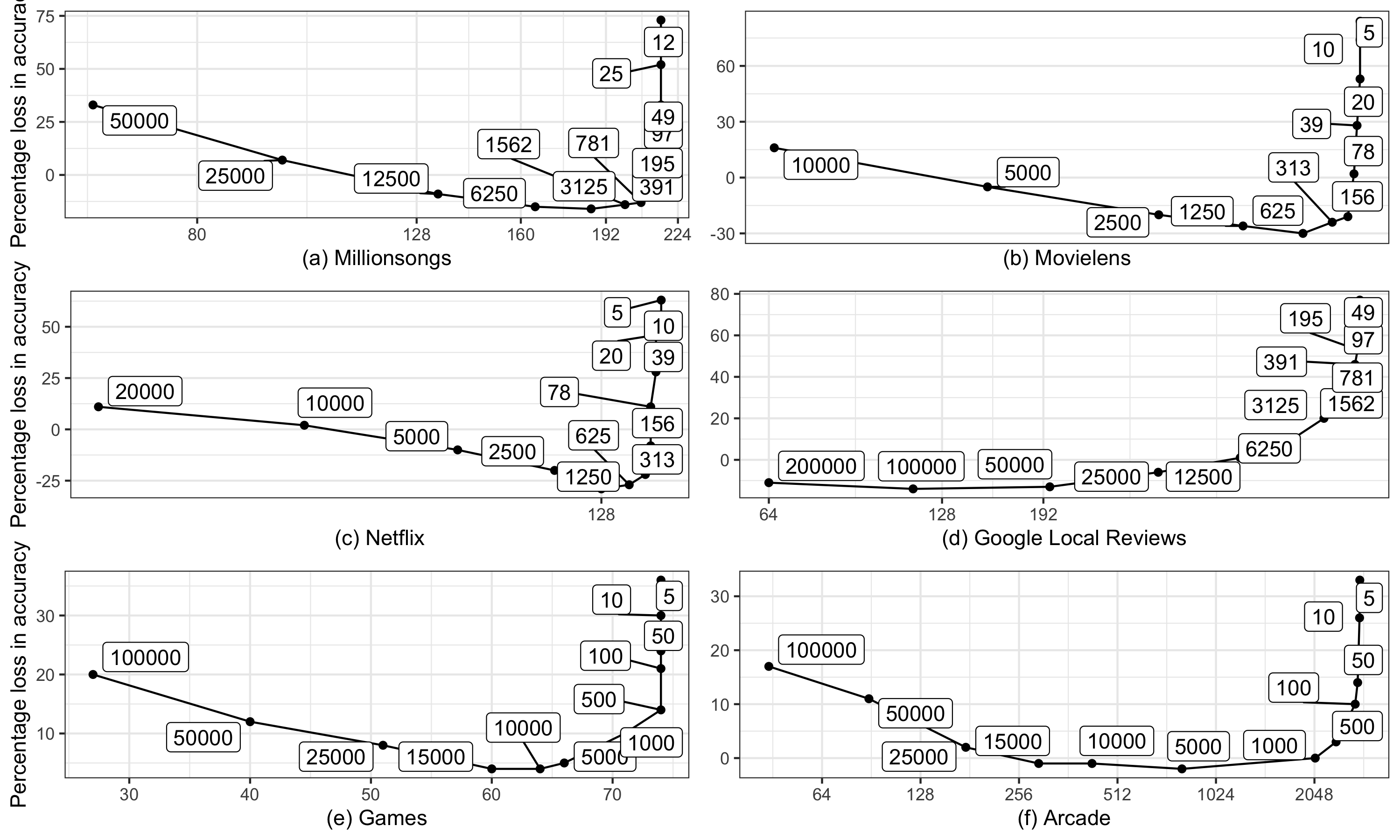} 
	\caption{Tuning the embedding size for given model size (= 20 MB). The x-axis denotes the embedding sizes and each datapoint is annotated with the corresponding number of embeddings.}
	\label{fig:plot9}
\end{figure*}

\textbf{Results.}  The plot~\ref{fig:plot9} shows the tradeoff between choosing a larger embedding size vs larger number of embeddings. It shows that for most use cases, including Millionsongs, Movielens, Netflix, Games and Arcade, the optimal number of embeddings for \texttt{MEmCom} is roughly 10x lower than its input vocabulary. Interestingly, this did not hold for the Google Local Reviews use case, where the distribution of reviews is more even across all entities due to geographical constraints.

\subsection{Lower precision} \label{sec:quantizationExperiments}

\textbf{Experimental Setup.} As mentioned in the section~\ref{sec:relatedWorks}, we can reduce the size of a model compressed via \texttt{MEmCom} by either reducing the floating point precision of weights and by sparsifying the weights~\cite{han2015learning, han2015deep_compression, Murray94enhancedmlp, Vanhoucke, DBLP:journals/corr/abs-1806-08342}. 
%As the gains due to sparsification will work irrespective of the , 
We leave the latter as a future work and focus on the effectiveness of quantization for model compression. We use the model described in the section~\ref{sec:hyperparams} and evaluate the loss in accuracy compared to a model that is compressed using \texttt{MEmCom} and whose weights are stored in single-point precision (i.e. 32 bits). By reducing the floating point precision by half, we reduce the model size by half. Hence, we use the floating point precision as x-axis to compare the effect of quantization across multiple datasets. We evaluate the results using CoreML\footnote{\url{https://developer.apple.com/machine-learning/core-ml/}} and with the quantization mode set to \texttt{linear}.

\textbf{Results.} As shown in the figure~\ref{fig:quantizationPlot}, all the datasets (except Google Local Review) have no loss in accuracy when the model is converted to half-point precision. The model trained on MovieLens can be compressed even further using 8-bit precision without any loss in accuracy. On all other datasets, the loss of accuracy is approximately 0.13\%  when using 8-bit precision. This implies that a typical recommender model can be compressed further by 4x using 8-bit precision with negligible loss in accuracy. However, the accuracy drops significantly if we quantize the model further for all the datasets. 

\begin{figure}
	\centering
	\includegraphics[width=0.47\textwidth]{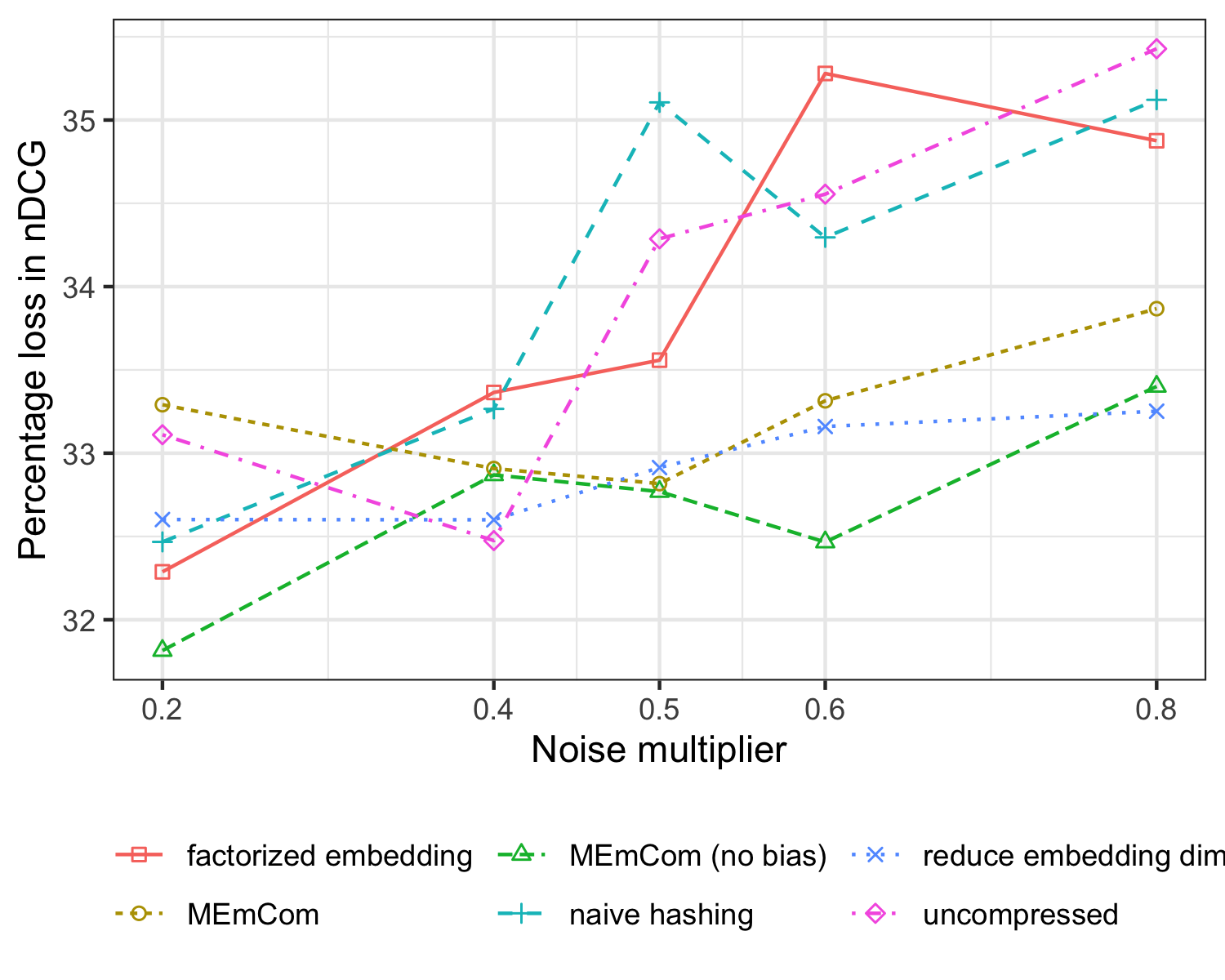}
	\caption{Evaluating privacy vs accuracy tradeoff: Arcade}
	\label{fig:plot4}
\end{figure}

\subsection{Robustness to noise for private federated learning} \label{sec:DPExperiment}

\textbf{Experimental Setup.}  Models deployed on-device have greater access to user data. 
This data is typically not sent to the server to protect user's privacy. To train such a model, one uses private federated learning, where the models are trained in a distributed manner across multiple devices. To guarantee user privacy, ``noise'' is introduced during training which
makes it difficult to infer whether any particular data points was used during training. 
However, this may degrade model accuracy compared to one trained without noise.
In this experiment, we focus on the tradeoff robustness of model compression techniques vs. noise.
We use an uncompressed model without noise as a baseline and evaluate percentage loss in nDCG 
for a model trained with a given noise multiplier.
We use the same dataset as described in the section~\ref{sec:rankingExperiment}.
Other than \texttt{reduce embedding dim}, we set the hyperparameters such that the compressed models were 51 MB in size.
The model compressed using \texttt{reduce embedding dim} was 102 MB in size.
To simulate differential private federated learning, we trained the models
with the R\'enyi Differential Privacy (RDP) framework~\cite{DBLP:journals/corr/Mironov17, AppleDifferentialPrivacy} 
using the TensorFlow Privacy library.
We use used global DP setup, constant l2$\_$norm$\_$clip. and set RDP's $\delta$ parameter to $\frac{1}{number \; of \; training \; points}$.
Since the TensorFlow Privacy library is only supported for TensorFlow 2.0 or greater, we used TensorFlow 2.4.0.

\textbf{Results.} The plot~\ref{fig:plot4} compares the relative accuracy (in our case nDCG) of different approaches
to that of an uncompressed model trained without noise for different noise multipliers.
It shows that our approach has lower loss in nDCG for a given noise multiplier and 
was more robust to noise that an uncompressed model and naive hashing.

\subsection{Practical sanity check that MEmCom produces unique embeddings}

\textbf{Experimental Setup.} On one model that was trained on the Arcade dataset using MEmCom, with an input embedding compression ratio of 40x, we examined the uniqueness of the embeddings that were produced. This experiment allowed us to test our claim that MEmCom is able to produce a unique embedding for each category.

\textbf{Results.} We found that a vanishingly small number of categories that shared an $\textbf{x}_{rem}$ embedding, as defined in Algorithm \ref{alg:ourApproachNoBias}, ended up with equal $\textbf{x}_{mult}$ multipliers. More precisely, a pair of multipliers sharing a common $\textbf{x}_{rem}$ embedding differed by greater than 0.00001 in more than 99.98\% of cases. This result validates the claim that MEmCom is able to produce a unique embedding for each category. 